\documentclass[letterpaper, 10 pt, conference]{ieeeconf}  

\IEEEoverridecommandlockouts                              

\usepackage{cite}
\usepackage{graphics} 
\usepackage{graphicx} 
\usepackage{amsmath} 
\usepackage{amssymb}  
\usepackage{amsthm}  
\usepackage{hyperref}

\usepackage{booktabs} 
\usepackage{xcolor}
\usepackage{colortbl}
\usepackage{multirow}
\usepackage{makecell}
\usepackage{balance}

\title{\LARGE \bf
Spectral Normalization for Lipschitz-Constrained Policies on Learning Humanoid Locomotion
}

\author{Jaeyong Shin$^{1}$, Woohyun Cha$^{1}$, Donghyeon Kim$^{1,2}$, Junhyeok Cha$^{1}$, and Jaeheung Park$^{3}$
\thanks{*This research was supported by Basic Science Research Program through the National Research Foundation of Korea(NRF) funded by the Ministry of Education(RS-2023-00274280)}
\thanks{$^{1}$Department of Intelligence and Information, Graduate School of Convergence Science and Technology, Seoul National University, Republic of Korea
\{{\tt\small jasonshin0537, woohyun321, kdh0429, threeman1}\}@snu.ac.kr}%
\thanks{$^{2}$1X Technologies. This work was conducted while the author was at Seoul National University}%
\thanks{$^{3}$Department of Intelligence and Information, Graduate School of Convergence Science and Technology, ASRI, AIIS, Seoul National University,
Republic of Korea, and Advanced Institute of Convergence Technology
(AICT), Suwon, Republic of Korea {\tt\small park73@snu.ac.kr}}%
}

\begin{document}
\theoremstyle{definition}
\newtheorem{definition}{Definition}

\maketitle
\thispagestyle{empty}
\pagestyle{empty}

\begin{abstract}

Reinforcement learning (RL) has shown great potential in training agile and adaptable controllers for legged robots, enabling them to learn complex locomotion behaviors directly from experience. However, policies trained in simulation often fail to transfer to real-world robots due to unrealistic assumptions such as infinite actuator bandwidth and the absence of torque limits. These conditions allow policies to rely on abrupt, high-frequency torque changes, which are infeasible for real actuators with finite bandwidth.

Traditional methods address this issue by penalizing aggressive motions through regularization rewards, such as joint velocities, accelerations, and energy consumption, but they require extensive hyperparameter tuning. Alternatively, Lipschitz-Constrained Policies (LCP) enforce finite bandwidth action control by penalizing policy gradients, but their reliance on gradient calculations introduces significant GPU memory overhead. To overcome this limitation, this work proposes \textit{Spectral Normalization} (SN) as an efficient replacement for enforcing Lipschitz continuity. By constraining the spectral norm of network weights, SN effectively limits high-frequency policy fluctuations while significantly reducing GPU memory usage. Experimental evaluations in both simulation and real-world humanoid robot show that SN achieves performance comparable to gradient penalty methods while enabling more efficient parallel training.

\end{abstract}

\section{INTRODUCTION}

Reinforcement learning (RL) has emerged as a powerful framework for developing locomotion policies, leading to significant advancements in legged robots. Recent studies have demonstrated that deep RL can produce agile and adaptable controllers, achieving impressive performance in simulation across a variety of bipedal locomotion tasks \cite{siekmann2021sim, duan2022sim, kim_tbasedrl, tang2024humanmimic}. However, a major challenge remains, \textit{sim-to-real transfer}. Policies trained in simulation often fail to generalize to real-world conditions due to discrepancies in dynamics, friction, and sensor noise. Especially, many simulation environments assume idealized conditions, such as infinite actuator bandwidth, allowing abrupt torque changes to be perfectly executed. In contrast, real robotic systems have finite control bandwidth, making it difficult for actuators to respond instantaneously to high-frequency control signals. Training under these unrealistic assumptions often leads to policies that rely on rapid, high-frequency torque changes, resulting in undesirable vibrations and oscillatory behavior when deployed on real hardware. This issue highlights the need for training techniques that explicitly account for real-world hardware limitation.

\begin{figure}
    \centering
    \includegraphics[width=\linewidth]{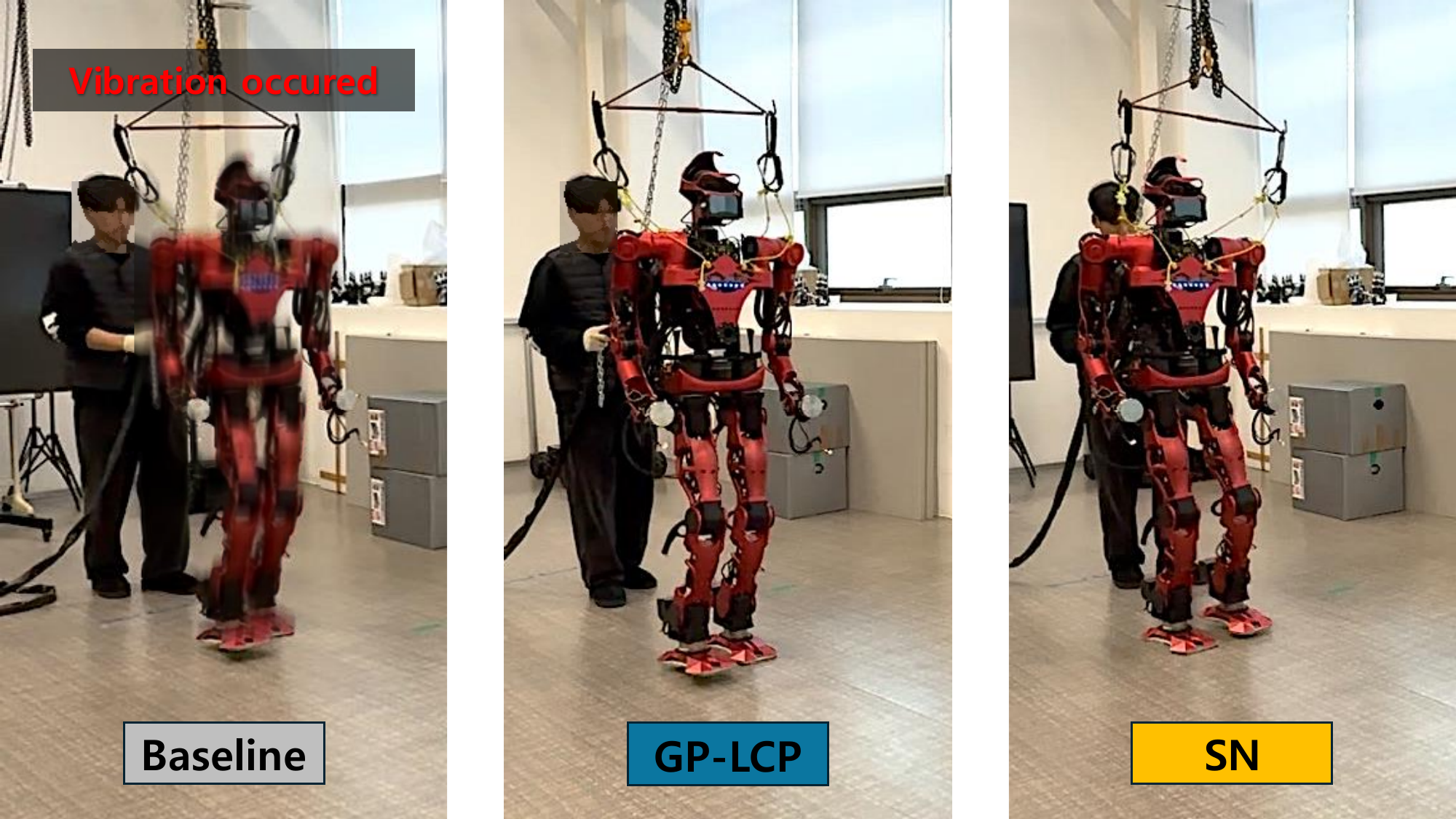}
    \caption{Comparison of policies trained using Baseline, Gradient Penalty-Based Lipschitz-Constrained Policy (GP-LCP), and Spectral Normalization (SN). The figure illustrates the regularization effects across methods. While the Baseline exhibits vibrations, GP-LCP stabilizes actions at the cost of increased computational overhead. SN achieves similar stability with lower computational overhead, demonstrating its effectiveness in enforcing finite bandwidth action control.}
    \label{fig:overview}
\end{figure}

Among the various factors contributing to the sim-to-real gap, this paper specifically focuses on the issue of vibration caused by high-frequency control variations. One critical aspect of addressing this problem, particularly in the presence of \textit{hardware limitations}, is limiting infinite bandwidth control output.
Previous approaches have introduced various strategies to enforce finite bandwidth actions in learned policies. A common method involves incorporating \textit{regularization rewards} into the objective function, such as regularizing joint velocities, accelerations, and energy consumption to suppress sudden control variations \cite{kim2024not, kim2024learning, fu2021minimizing}. While effective, regularization rewards introduce additional hyperparameters and require careful tuning to balance task performance and motion quality.

A more principled alternative is to enforce Lipschitz continuity in the policy network, ensuring that small changes in the input state produce proportionally small changes in the output action \cite{chen2024lcp}. Prior work has shown that incorporating gradient penalties in policy updates effectively constrains the rate of change in learned policies, leading to finite bandwidth action control. However, while gradient penalties provide a viable approach to limiting high-frequency components in actions, their substantial GPU memory requirements for policy gradient computation remain a critical limitation in practical applications. In large-scale reinforcement learning environments such as Isaac Gym \cite{makoviychuk2021isaac}, where thousands of parallel simulations are run to accelerate policy learning, this increased memory overhead can be problematic. Due to limited GPU memory, computing gradient penalties may require reducing the number of simulated environments, leading to slower training convergence and potential performance degradation, making gradient penalty-based LCP methods less practical.

To address this computational overhead, this work proposes applying spectral normalization (SN) to enforce Lipschitz continuity as a more effective approach to gradient penalties. Rather than explicitly penalizing policy gradients, SN constrains the spectral norm of each weight matrix, effectively bounding the policy’s Lipschitz constant without requiring costly gradient computations. This allows policies to enforce finite bandwidth actions with significantly lower GPU memory usage, enabling larger-scale parallel training in Isaac Gym without sacrificing computational efficiency. 
Experimental evaluations, as shown in \autoref{fig:overview}, conducted both in simulation and on a real robot platform, demonstrate that this method not only achieves comparable control stability to existing Lipschitz-constrained policies while reducing GPU memory consumption but also enhances sim-to-real transfer by considering hardware limitations such as actuator bandwidth and control constraints.

The remainder of this paper is organized as follows. \autoref{sec:related_work} reviews existing approaches to enforcing finite bandwidth action control in RL and discusses their limitations. \autoref{sec:background} provides the theoretical background on Lipschitz continuity and spectral normalization. \autoref{sec:methods} details the proposed approach, introducing spectral normalization as a better approach than gradient penalty-based regularization for limiting high-frequency components in policy outputs. \autoref{sec:training_setup} outlines the training environment, network architecture, and hyperparameter settings used in our experiments. \autoref{sec:experiments} presents quantitative and qualitative evaluations of the proposed method in both simulation and real-world settings. Finally, \autoref{sec:conclusions} summarizes key findings and discusses potential directions for future research.

\section{Related Work}
\label{sec:related_work}
Recent advances in reinforcement learning have demonstrated significant potential in developing robust controllers for locomotion tasks. In particular, policies learned via deep reinforcement learning have achieved impressive performance on simulated robotic platforms, suggesting a promising avenue for robust, agile, and adaptable locomotion \cite{cheng2024extreme, kumar2021rma, lai2023sim, kumar2022adapting, radosavovic2024real}. However, the challenge of sim-to-real transfer has become increasingly critical, as many simulated environments assume idealized conditions, such as infinite actuator bandwidth, that do not hold in real-world robotics. This discrepancy requires the development of training methods that account for the physical limitations of hardware.

\subsubsection{Regularization Reward}
One common strategy for promoting finite bandwidth action control in learned policies is the incorporation of regularization rewards into the training objective. These additional penalty terms typically discourage abrupt changes by penalizing large joint velocities, accelerations, and energy consumption \cite{he2024learning, he2024omnih2o, fu2024humanplus, liu2024visual, kim_tbasedrl, fu2021minimizing, gu2024humanoid}. Although effective at mitigating sudden or jittery motions, such rewards introduce additional hyperparameters that must be carefully tuned to balance task performance and motion quality.

\subsubsection{Gradient Penalty-based Lipschitz-Constrained Policy (GP-LCP)}
An alternative approach is to incorporate a penalty directly into the objective function that discourages large changes in the policy output with respect to variations in the input. \textit{Chen et al.} \cite{chen2024lcp} propose a gradient penalty that constrains the policy’s rate of change by penalizing large gradients. This constraint ensures that even substantial input variations result in gradual and controlled action. The strength of this constraint is controlled by a hyperparameter, allowing trade-offs between stability and responsiveness based on task requirements. While this method effectively eliminates unwanted high-frequency components in policy output, it significantly increases GPU memory usage due to the additional gradient computations required during training.

\subsubsection{Spectral Normalization}
Rather than applying an explicit gradient penalty, this work employs spectral normalization to constrain the Lipschitz constant of the policies. 
\textit{Miyato et al.} \cite{snforgan} introduced spectral normalization to enforce a bounded rate of change in the discriminator network of generative adversarial networks (GANs), improving training stability by preventing excessive gradients.
Spectral normalization can be adopted to regulate the Lipschitz constant of neural networks. By normalizing weight matrices based on their largest singular value, spectral normalization directly constrains the maximum rate of change of the network output with respect to its input. Integrating spectral normalization into policy networks provides a computationally efficient means of limiting high-frequency action variations, reducing GPU memory overhead while maintaining constrained bandwidth actions.

\section{Background}
\label{sec:background}
\subsection{Lipschitz Continuity}
Lipschitz continuity is a property that restricts the rate at which a function can change. It is a useful property for evaluating the smoothness of a function. Formally, it can be defined as follows:

\begin{definition}
A function \( f: \mathbb{R}^n \to \mathbb{R}^m \) is said to be \textit{Lipschitz continuous} if there exists a constant \( L \geq 0 \) such that for all \( x, y \in \mathbb{R}^n \), the following inequality holds:
\begin{equation}
    \| f(x) - f(y) \| \leq L \| x - y \|.
\end{equation}
The constant \( L \) is referred to as the \textit{Lipschitz constant} \cite{o2006metric}.
\end{definition}

\subsection{Reinforcement Learning}
To model a robot control problem, a discrete-time Markov Decision Process (MDP) is frequently used. At each time step $t$, the agent observes the state space $s_t$ which is the input of policy $\pi$ and samples an action $a_t$ from the policy $a_t \sim \pi(a_t|s_t)$ based on the observed state. The agent then applies the action, which transit the robot to new state $s_{t+1}$, and a reward $r_t = r(s_t,a_t,s_{t+1})$. The objective of the agent is to learn a policy that maximizes the expected discounted return $J(\pi_{\theta})$, 
\begin{equation} \label{mdp}
J(\pi_{\theta}) = E_{\tau \sim p(\tau|\pi_{\theta})} \left[\sum_{t=0}^{T-1} \gamma^tr_t \right],
\end{equation}
where $\tau$ is the trajectory under a policy $\pi_{\theta}$. $T$ represents the finite-horizon to measure cumulative discounted reward and $\gamma$ is a discount factor. 

\subsection{Biped Robot TOCABI}
In this study, we utilize the bipedal robot TOCABI for hardware validation. TOCABI is a humanoid robot with a height of 1.8 meters and a weight of 100 kilograms \cite{schwartz2022design}. Its design closely mirrors human proportions, featuring 12 actuated joints in the lower body and 21 actuated joints in the upper body. Due to its substantial weight, high reduction ratio gears of 100:1 are employed. The robot's current-controlled servo drives are managed through an EtherCAT interface, which connects to a real-time control computer and operates each joint motor at a frequency of 2 kHz.

In this work, we implement RL with torque-based control to TOCABI on 12 actuated joints in the lower body, while the upper body was maintained in a stable posture using position control.

\begin{figure}
    \centering
    \includegraphics[width=\linewidth]{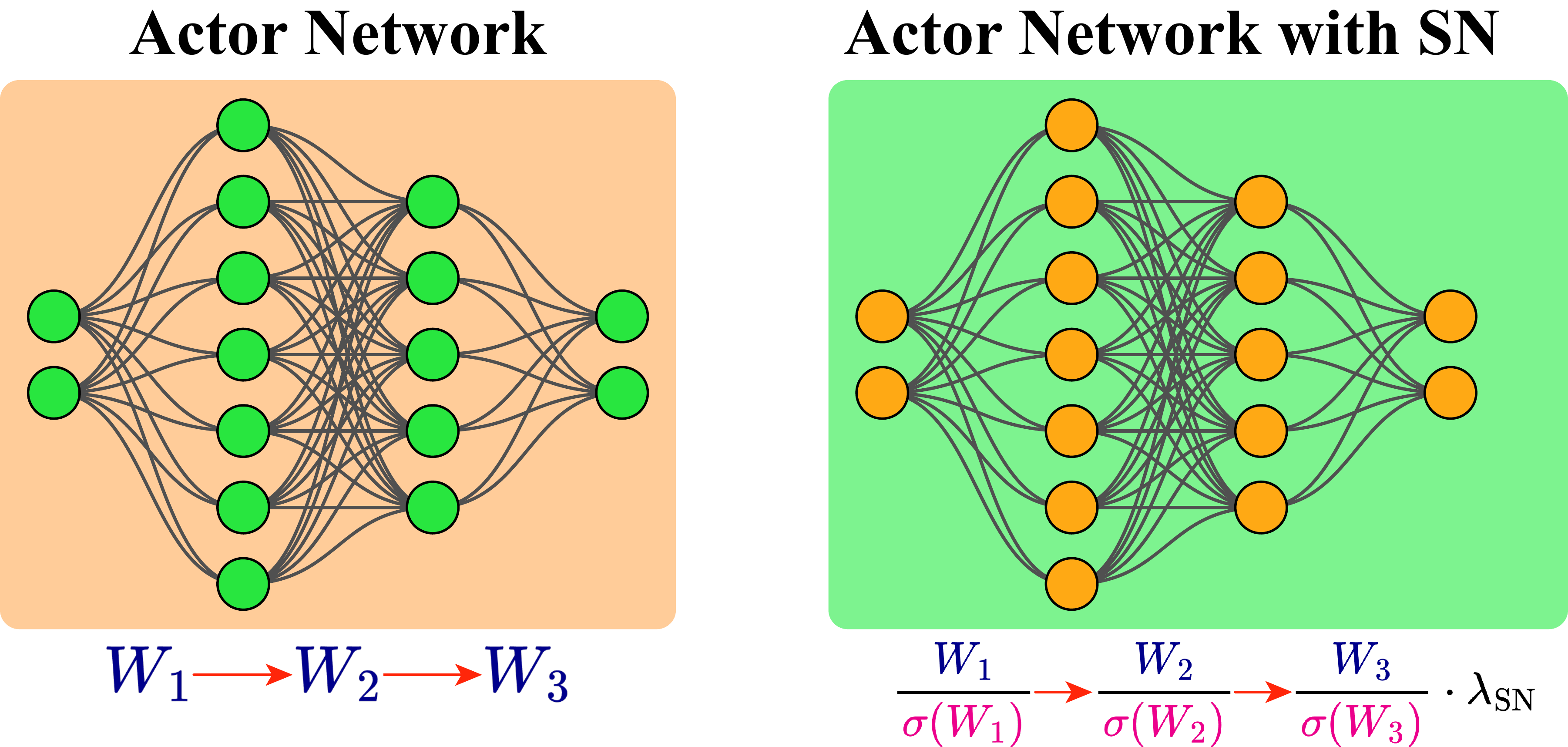}
    \caption{Comparison between a standard actor network and an actor network with SN. In the SN-based network, each weight matrix  $W_l$  is divided by its largest singular value  $\sigma(W_l)$, effectively constraining the Lipschitz constant of the network. Unlike conventional networks, where weights can grow arbitrarily, SN ensures bounded transformations at each layer.}
    \label{fig:overview-sn}
\end{figure}
\section{Methods}
\label{sec:methods}
\subsection{Spectral Normalization for Lipschitz-Constrained Policies}
\subsubsection{Motivation for Spectral Normalization}
When learning policies for humanoid locomotion, one common issue is the emergence of jittery or oscillatory movements. To address this issue, many studies have relied on hand-crafted regularization rewards, such as penalizing rapid changes in joint positions or action commands. However, while these reward terms can reduce jitter, they also introduce additional hyperparameters and tuning overhead.

Recent work on Lipschitz-Constrained Policies (LCP) demonstrates that ensuring that the policy network is Lipschitz continuous effectively promotes finite bandwidth action outputs without resorting to extensive reward shaping. Specifically, GP-LCP methods show that if the policy is Lipschitz bounded, small perturbations in the input lead to correspondingly small changes in the output, thereby reducing undesirable vibration on actuators. 
Despite these benefits, GP-LCP technique relies on gradient penalties that significantly increase GPU memory usage during training. In practice, this can limit the number of parallel training environments, reducing data collection efficiency and slowing convergence. This trade-off makes GP-LCP methods less practical for large-scale reinforcement learning.

\subsubsection{Proposed Approach}
In this work, we propose to enforce Lipschitz continuity using spectral normalization instead of gradient penalties. Originally introduced in the context of Generative Adversarial Networks (GANs) \cite{snforgan}, spectral normalization constrains the spectral norm of each weight matrix in the neural network, thus bounding the Lipschitz constant of the network. Since it does not require computing and penalizing gradient norms directly, spectral normalization alleviates the high GPU memory demands typically associated with gradient-penalty-based methods.

\subsubsection{Spectral Normalization}
Spectral normalization (SN) operates by rescaling each weight matrix $W$ based on its largest singular value, $\sigma(W)$, thereby controlling the Lipschitz constant of the resulting function. In particular, for a weight matrix $W$, we estimate its spectral norm $\sigma(W)$ and then normalize the weights:
\begin{equation}
\hat{W} \;=\; \frac{W}{\sigma(W)},
\end{equation}
which ensures that Lipschitz norm $\|\hat{W}\|_{\text{Lip}}\leq 1$. In the actor network $\mu = f(s,\theta)$, $f$ is a neural network of the following form with input $s$:
\begin{equation}
    f(s, \theta) = W_{L+1} a_L \big(W_L a_{L-1} (W_{L-1} \dots a_1(W_1 s) \dots ) \big),
\end{equation}
where $\theta := \{ W_1, \dots, W_L, W_{L+1} \}$ is the set of learning parameters, $
W_l \in \mathbb{R}^{d_l \times d_{l-1}}, W_{L+1} \in \mathbb{R}^{1 \times d_L},$ and $ a_l$ is an element-wise non-linear activation function. Let $\mathbf{h}_l$ denote the hidden representation at layer $l$, with $\mathbf{h}_0 = s$. \textit{Miyato et al.} \cite{snforgan} showed the following bound on $\|f\|_{\text{Lip}}$:

\begin{equation}
\begin{aligned}
    \|f\|_{\text{Lip}} &\leq \|( \mathbf{h}_L \mapsto W_{L+1} \mathbf{h}_L )\|_{\text{Lip}} \\
    &\quad \cdot \|a_L\|_{\text{Lip}} \cdot \|( \mathbf{h}_{L-1} \mapsto W_L \mathbf{h}_{L-1} )\|_{\text{Lip}} \\
    &\quad \cdots \|a_1\|_{\text{Lip}} \cdot \|( \mathbf{h}_0 \mapsto W_1 \mathbf{h}_0 )\|_{\text{Lip}} \\
    &= \prod_{l=1}^{L+1} \|( \mathbf{h}_{l-1} \mapsto W_l \mathbf{h}_{l-1} )\|_{\text{Lip}} \\
    &= \prod_{l=1}^{L+1} \sigma(W_l),
\end{aligned}
\end{equation}
assuming each activation function $a_l$ is 1-Lipschitz. By constraining $\sigma(W^l) \le 1$ for all layers, f is 1-Lipschitz. This property promotes finite bandwidth action outputs and reduces jitter in locomotion tasks since small changes in the state s lead to proportionally small changes in the actions $\mu$.

\subsubsection{Lipschitz Constrained Policy with SN}
Although normalizing each layer in the actor network ensures that the resulting action is 1-Lipschitz in $\|\cdot\|$-norm, LCP requires the probability distribution over actions itself to be robust against small perturbations in the state. Specifically, for policy-gradient methods, the key quantity of interest is not just the raw action output, but $\log \pi(a \mid s)$. Consequently, a constrained policy optimization problem can be formulated to enforce Lipschitz continuity through a gradient constraint:
\begin{equation}
\begin{aligned}
\label{eq:4}
    \max_{\pi} \quad & J(\pi) \\
    \text{s.t.} \quad & \max_{s,a} \big[\| \nabla_{s}\,\log\,\pi(a|s)\|^2\big] \le K^2
\end{aligned}
\end{equation}
where $K$ is a constant and $J(\pi)$ is the RL objective described in \eqref{mdp}. The term $\log\,\pi(a|s)$ denotes the log-likelihood of selecting action $a$ in state $s$, a key quantity in policy gradient methods. The following derivation demonstrates how spectral normalization can help ensure the gradient satisfies this Lipschitz constraint:

\begin{equation}
\label{eq7}
\begin{split}
    &\max_{s,a} \big[\| \nabla_{s}\,\log\,\pi(a|s)\|^2\big] \\
    &= \max_{s,a} \Bigg[\bigg\| \nabla_s \Big[
        -\frac{1}{2} \bigl(\frac{\bigl(a-\mu(s)\bigr)^2}{\sigma^2} 
        + \log(2\pi\sigma^2)\bigr)\Big] \bigg\|^2\Bigg] \\
    &= \max_{s,a} \bigg[\frac{\|a-\mu(s)\|^2}{\sigma^4} \; \|\nabla_s\,\mu(s) \|^2 \bigg] \\
    &\lesssim \frac{(2\sigma)^2}{\sigma^4} \cdot \|\nabla_s\,\mu(s) \|^2 \\
    &= \frac{4}{\sigma^2}\|\nabla_s\,\mu(s) \|^2,
\end{split}
\end{equation}

In the derivation above, the standard deviation $\sigma$ remains fixed during optimization, and $\mu(s)$ is produced by the actor network whose weight matrices have been normalized via spectral normalization. Under a gaussian policy $\pi(a|s)\;= \mathcal{N}(\mu(s),\sigma^2)$, approximately 95\% of the samples $a$ will lie within two standard deviations of the mean. Accordingly, the maximum squared gradient may be approximated as 
\begin{equation}
    \max_{s,a} \big[\| \nabla_{s}\,\log\,\pi(a|s)\|^2\big] \approx \frac{4}{\sigma^2}\|\nabla_s\,\mu(s) \|^2,
\end{equation}

Since spectral normalization enforces an upper bound on the Lipschitz constant of each layer in the actor network, it effectively controls $\|\nabla_s \mu(s)\|$. By bounding $\|\nabla_s \mu(s)\|$, spectral normalization helps ensure that the gradient $\nabla_s \log \pi(a|s)$ adheres to the desired Lipschitz constraint. 
Hence, the constraint in \eqref{eq:4} implies that $K \approx 2/\sigma$ if the weight matrices of the actor networks are spectral normalized.

\begin{figure}
    \centering
    \includegraphics[width=0.8\linewidth]{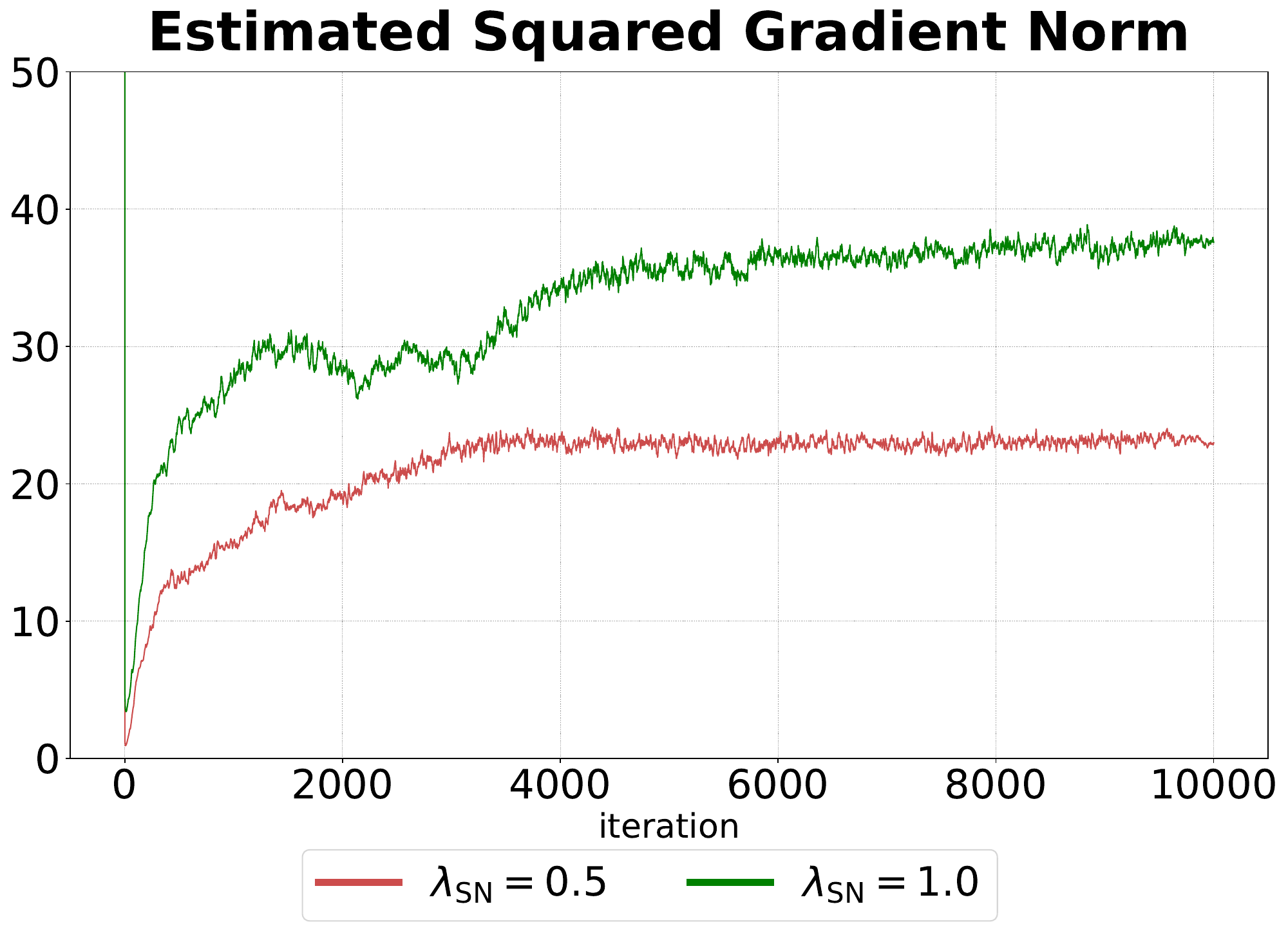}
    \caption{Estimated squared gradient norm during training process}
    \label{fig:gploss}
\end{figure}
Furthermore, to allow flexible control over the Lipschitz constant $K$, an additional hyperparameter, referred to here as the \emph{SN coefficient} $(\lambda_{\text{SN}})$, was introduced in the final layer of the actor network. \autoref{fig:overview-sn} provides a simplified comparison between the conventional actor network and the proposed actor network with SN applied. By scaling the gradient in this layer, practitioners can fine-tune the policy’s overall $K$-Lipschitz behavior as required.

To verify that the policy indeed satisfies the $2\lambda_\text{SN}/\sigma$-Lipschitz continuity in the state space, the magnitudes of the gradient norm were estimated during training. The standard deviation $\sigma$ was fixed at 0.2, throughout the training. SN was applied to every layer of the actor network, and the final layer’s weights were multiplied by a selected SN coefficient, $\lambda_{\text{SN}}$. In the case of $\lambda_{\text{SN}} = 1.0$, the estimated squared gradient norm must remain below 100 with 95\% probability. For $\lambda_{\text{SN}} = 0.5$, it must stay below 25 with the same probability. \autoref{fig:gploss} confirms that both conditions are satisfied throughout training, indicating that the policy complies with the intended $2\lambda_\text{SN}/\sigma$-Lipschitz bound in the state space for the specified values of $\lambda_{\text{SN}}$. These results demonstrate that the SN-based policy indeed meets the desired Lipschitz constraint in practice.

\subsubsection{Fast Approximation of SN}
To perform spectral normalization, the largest singular value of a weight matrix must be computed. Direct computation via singular value decomposition (SVD) is computationally expensive, particularly when it must be performed repeatedly during training and deploying on the real-world. To address this issue, the power iteration method is employed to efficiently estimate the largest singular value \cite{yoshida2017spectral}. The method and algorithm closely follow those presented in \cite{snforgan}. Notably, even a single iteration of the power iteration algorithm yields a relatively accurate approximation, making it a practical and scalable solution for spectral normalization in actor network.

\section{Training Setup}
\label{sec:training_setup}
To evaluate the effectiveness of Spectral Normalization on actor network, TOCABI was trained based on the Adversarial Motion Prior (AMP) framework \cite{peng2021amp}, where the task is to imitate reference motion data and to track base velocity based on the velocity command inputs. 

\subsection{Observations and Actions}
To ensure the policy can effectively follow the reward signals, selecting the right input features is crucial. The observation space $\mathcal{O} \subset \mathbb{R}^{49}$ is composed of the following elements:
\begin{enumerate}
    \item Base height $h \in \mathbb{R}$ 
    \item Base Euler angle $\alpha, \beta, \gamma \in \mathbb{R}^3$ 
    \item Base linear and angular velocity $ \nu, \omega \in \mathbb{R}^6$, expressed in base local coordinate frame 
    \item Velocity command (x, yaw)$\in \mathbb{R}^2$
    \item Joint position $q \in \mathbb{R}^{12}$
    \item Joint velocity $\dot{q} \in \mathbb{R}^{12}$ 
    \item Previous action $\in \mathbb{R}^{12}$
\end{enumerate}
The action space $A \in \mathbb{R}^{12}$ is composed of 12 actions to generate joint torque commands and the RL policy is utilized to directly produce the torque commands at a frequency of 250Hz.

\subsection{Imitation Rewards}
This work adopts the AMP framework \cite{peng2021amp} to guide bipedal locomotion. AMP trains a discriminator $D(s, s')$ to distinguish between state-transition pairs drawn from a reference dataset $M$ and those generated by the policy. In practice, $D$ is updated to assign a score of 1 to transitions from the reference dataset and -1 to transitions produced by the policy. The policy then receives a \emph{style reward}:
\begin{equation}
    r_{style}(s_t, s_{t+1}) = \max\Bigl[0,\;1 - 0.25\bigl(D(s_t,s_{t+1}) - 1\bigr)^2\Bigr],
\end{equation}
which is included in the RL objective from \eqref{mdp}. The input $s$ to $D$ includes features that capture the robot's configuration and movement (e.g., base orientation, joint position and velocity, and local foot positions).

\subsection{Task Specific Rewards}
While the style reward $r_{style}$ ensures that the policy generates motions that are visually consistent with the reference data, task-specific rewards drive the agent to fulfill the functional objectives of the task. 
Our agents operate within a command-conditioned framework, where they interact with the environment by following specific commands during training. The local linear velocity command along the x-axis and the angular velocity command around the yaw axis are constrained within the range [-0.5, 0.5]. To ensure that the robot's pelvis link accurately tracks these commands, task reward is specified as,
\begin{equation}
    r_{task} = \alpha \cdot exp(-\beta \lVert v_t^{cmd} - v_t \rVert_2^2)
\end{equation}
where $\alpha$ and $\beta$ are tunning parameters for training.

\subsection{Training Details}
\begin{table}
    \centering
    \caption{Sim-to-Real Randomization}
    \resizebox{\columnwidth}{!}{%
        \begin{tabular}{clcc}
            \toprule
            \rowcolor{gray!20} 
            Contents & Control Parameters & Range & Units \\
            \midrule
            \multirow{4}{*}[-1em]{\makecell[c]{Domain \\ Randomization}}  
                      & Mass              & [0.8, 1.2] $\times m_{default}$    & kg \\ \cmidrule(lr){2-4}
                      & Joint damping     & [0.5, 2.5]    & Nm \( \cdot \) s/rad\\ \cmidrule(lr){2-4}
                      & Joint armature    & [0.8, 1.2] $\times \; I_{default}$    & kg\( \cdot \)m\textsuperscript{2}\\ \cmidrule(lr){2-4}
                      & Motor constant    & [0.8, 1.2] $\times \; c_{default}$   & - \\ 
            \midrule
            \multirow{5}{*}[-1.2em]{\makecell[c]{Noise, \\ Bias, \\ Delay}}     
                      & \makecell[l]{Joint position,\\velocity noise} & $N \sim (0, 0.0005)$ & - \\ \cmidrule(lr){2-4}
                      & Base velocity noise            & [-0.025, 0.025]     & rad/s \\ \cmidrule(lr){2-4}
                      & Joint position bias            & [-0.0314, 0.0314]    & rad \\ \cmidrule(lr){2-4}
                      & Base orientation bias          & [-0.02, 0.02]       & rad \\ \cmidrule(lr){2-4}
                      & Action delay                   & [0.002, 0.01]       & s \\ 
            \bottomrule
        \end{tabular}%
    }
    \label{tab:domain_randomization}
\end{table}

The proposed method was trained using Proximal Policy Optimization (PPO) \cite{schulman2017proximal}. The total loss function consisted of the standard PPO objective combined with an AMP loss from \cite{peng2021amp}, formulated as $L_{\text{total}} = L_{\text{PPO}} + L_{\text{AMP}}$. The discriminator was modeled as a fully connected MLP with two hidden layers of 256 neurons, using ReLU activations.
The learning rate was set to $1 \times 10^{-4}$ with a linear scheduler decreasing to $1 \times 10^{-6}$. Both the policy and value networks were modeled as fully connected MLPs with two hidden layers of 512 neurons, using ReLU activations. Spectral normalization was applied to each layer of the policy network to constrain its Lipschitz constant. The weights of the final layer were multiplied by a selected SN coefficient, $\lambda_{\text{SN}}$. The action distribution was parameterized as a Gaussian with a fixed standard deviation $\sigma=0.2$.

\begin{table*}[t]
    \centering
    \caption{Ablation Studies. All policies are trained with three random seeds and tested in 1000 environments for 2500 steps, corresponding to 10 seconds clock time.}
    \resizebox{2\columnwidth}{!}{%
        \begin{tabular}{l c c c c c}
        \toprule
        \multicolumn{6}{l}{\cellcolor{gray!20}\textbf{(a) Ablation on Regularization Methods}} \\
        \midrule
        Method & Joint Velocity & Joint Acceleration & Torque Difference & Energy & Task Return \\
        \midrule
        Baseline           & 1.7341 $\pm$ 1.0457 & 0.2186 $\pm$ 0.1375 & 27.36 $\pm$ 36.66 & 35.82 $\pm$ 90.1742 & \textbf{0.0037 $\pm$ 0.0089} \\
        Regularization Reward  & \textbf{1.7133 $\pm$ 1.0669} & 0.2138 $\pm$ 0.1311 & 18.29 $\pm$ 19.89 & 37.99 $\pm$ 111.88 & 0.0040 $\pm$ 0.0087 \\
        GP-LCP                & 1.7470 $\pm$ 1.0926 & 0.2084 $\pm$ 0.1363 & 9.80 $\pm$ 12.58 & \textbf{31.86 $\pm$ 85.00} & 0.0042 $\pm$ 0.0099 \\
        SN $\lambda_{\text{SN}}=0.2$ (Ours) & 1.8134 $\pm$ 1.1473 & \textbf{0.2032 $\pm$ 0.1261} & \textbf{9.25 $\pm$ 8.48} & 32.37 $\pm$ 86.28 & 0.0039 $\pm$ 0.0098 \\
        \midrule
        \multicolumn{6}{l}{\cellcolor{gray!20}\textbf{(b) Ablation on SN coefficient ($\lambda_{\text{SN}}$)}} \\
        \midrule
        Method & Joint Velocity & Joint Acceleration & Torque Difference & Energy & Task Return \\
        \midrule
        SN $\lambda_{\text{SN}}=1.0$ & 1.8927 $\pm$ 1.0518 & 0.2117 $\pm$ 0.1336 & 21.29 $\pm$ 20.78 & 37.39 $\pm$ 96.19 & \textbf{0.0035 $\pm$ 0.0094} \\
        SN $\lambda_{\text{SN}}=0.5$ & \textbf{1.7756 $\pm$ 1.0207} & 0.2047 $\pm$ 0.1282 & 19.49 $\pm$ 25.76 & 32.76 $\pm$ 74.66 & 0.0037 $\pm$ 0.0091 \\
        SN $\lambda_{\text{SN}}=0.2$ (Ours) & 1.8134 $\pm$ 1.1473 & \textbf{0.2032 $\pm$ 0.1261} & \textbf{9.25 $\pm$ 8.48} & \textbf{32.37 $\pm$ 86.28} & 0.0039 $\pm$ 0.0098 \\
        SN $\lambda_{\text{SN}}=0.1$ & \multicolumn{5}{c}{\cellcolor{red!10}\textbf{Failed to learn}} \\
        \midrule
        \bottomrule
        \end{tabular}%
    }
    \label{tab:ablation_studies}
\end{table*}
To mitigate the sim-to-real gap and enhance policy robustness, randomization techniques were also incorporated during training. As detailed in \autoref{tab:domain_randomization}, domain randomization, noise, bias, and delay were applied. 
Training was conducted in Isaac Gym \cite{makoviychuk2021isaac} with 4096 parallel environments. Each episode lasted 8000 timesteps, with actions applied at 250 Hz. A total of 131072 samples were collected per policy update, for a total of 20000 policy updates.
Experiments were performed on an RTX 4090 GPU, with all computations executed on GPU to accelerate training. Average training time took approximately 4 to 5 hours.

\section{Experiments}
\label{sec:experiments}
This section examines the impact of SN on enforcing finite bandwidth action outputs by comparing it with other control regularization techniques. Specifically, the following approaches are evaluated:
\begin{itemize}
    \item \textbf{Baseline (No Regularization):} 
    A policy without any regularization reward is included to illustrate the importance of finite bandwidth action outputs for sim-to-real transfer.
    \item \textbf{Regularization Reward:}
    This approach penalizes large or abrupt actions in different ways to enforce constrained bandwidth actions. Four types of regularization are considered:
    \begin{enumerate}
        \item \textit{Joint Velocity Regularization}
        \item \textit{Joint Acceleration Regularization}
        \item \textit{Torque Regularization}
        \item \textit{Torque Difference Regularization}
    \end{enumerate}
    \item \textbf{Gradient Penalty-based LCP (GP-LCP):}
    By penalizing large policy gradients, this technique limits how quickly the network output can change in response to input variations, thereby mitigating high-frequency oscillations.
\end{itemize}

To evaluate the effectiveness of each method, several action regularization metrics are used, including \textit{Joint Velocity, Joint Acceleration, Torque Difference}, and \textit{Mean Energy}. 
\textit{Joint Velocity} (rad/s) measures the overall magnitude of joint movements at each timestep by computing the L2 norm of velocity values across leg joints.
Similarly, \textit{Joint Acceleration} (rad/s\textsuperscript{2}) represents how quickly joint velocities change and is computed in the same manner. \textit{Torque Difference} (N\( \cdot \)m/s) quantifies variations in control torques between consecutive timesteps by calculating the differences in torque across leg joints in the same manner. Large torque difference indicates abrupt force changes, which can introduce vibrations and negatively impact control stability. 
For convenience, Torque Difference and Joint Acceleration were computed without dividing by the fixed timestep  dt = 0.004, as the constant value does not affect relative comparisons between methods.
\textit{Mean Energy} (W) is computed as the sum of the product of joint torque commands and joint velocity across leg joints. 
In addition, \textit{mean task return}, which quantifies the base linear and angular velocity tracking error, is also examined. The error is computed as the squared Euclidean distance between the desired and actual velocities in both the linear \( x \)-direction and yaw angular velocity.
\begin{table*}
    \centering
    \caption{Real Robot Deployment. All policies are trained with three random seeds and tested three types of commands for 1250 steps, corresponding to 5 seconds clock time.}
    \resizebox{2\columnwidth}{!}{%
        \begin{tabular}{l c c c c c}
        \toprule
        \multicolumn{6}{l}{\textbf{\cellcolor{gray!20}(a) Forward Walking $v_x = 0.25, w_{yaw} = 0.0$}} \\
        \midrule
        Method & Joint Velocity & Joint Acceleration & Torque Difference & Energy & Task Return \\
        \midrule
        Baseline           & \multicolumn{5}{c}{\cellcolor{red!10}\textbf{Failed to walk, vibration occured}}\\
        Regularization Reward  & 1.9720 $\pm$ 0.9612 & 0.0271 $\pm$ 0.0117 & 13.74 $\pm$ 8.13 & 75.11 $\pm$ 114.84 & 0.0181 $\pm$ 0.0222 \\
        GP-LCP                & 1.7551 $\pm$ 1.0173 & 0.0222 $\pm$ 0.0160 & 16.30 $\pm$ 55.12 & 55.12 $\pm$ 87.32 & \textbf{0.0155 $\pm$ 0.0261} \\
        SN $\lambda_{\text{SN}}=0.2$ (Ours) & \textbf{1.3663 $\pm$ 0.9203} & \textbf{0.0206 $\pm$ 0.0149} & \textbf{7.43 $\pm$ 12.19} & \textbf{47.96 $\pm$ 79.22} & 0.0315 $\pm$ 0.0456 \\
        \midrule
        \multicolumn{6}{l}{\cellcolor{gray!20}\textbf{(b) Rotational Walking $v_x = 0.0, w_{yaw} = 0.4$}} \\
        \midrule
        Method & Joint Velocity & Joint Acceleration & Torque Difference & Energy & Task Return \\
        \midrule
        Baseline           & \multicolumn{5}{c}{\cellcolor{red!10}\textbf{Failed to walk, vibration occured}}\\
        Regularization Reward  & 1.1520 $\pm$ 0.5746 & 0.0209 $\pm$ 0.0134 & 10.74 $\pm$ 8.27 & 34.75 $\pm$ 34.86 & 0.0120 $\pm$ 0.0225 \\
        GP-LCP                & 1.0993 $\pm$ 0.4369 & 0.0124 $\pm$ 0.0047 & 4.48 $\pm$ 3.86 & 28.99 $\pm$ 18.70 & \textbf{0.0062 $\pm$ 0.0178} \\
        SN $\lambda_{\text{SN}}=0.2$ (Ours) & \textbf{0.6583 $\pm$ 0.5799} & \textbf{0.0095 $\pm$ 0.0086} & \textbf{3.46 $\pm$ 4.29} & \textbf{16.15 $\pm$ 18.76} & 0.0154 $\pm$ 0.0234 \\
        \midrule
        \multicolumn{6}{l}{\cellcolor{gray!20}\textbf{(c) Forward + Rotational Walking $v_x = 0.1, w_{yaw} = 0.2$}} \\
        \midrule
        Method & Joint Velocity & Joint Acceleration & Torque Difference & Energy & Task Return \\
        \midrule
        Baseline           & \multicolumn{5}{c}{\cellcolor{red!10}\textbf{Failed to walk, vibration occured}}\\
        Regularization Reward  & 1.3313 $\pm$ 0.8651 & 0.0249 $\pm$ 0.0214 & 12.25 $\pm$ 10.71 & 42.18 $\pm$ 69.10 & \textbf{0.0229 $\pm$ 0.0293} \\
        GP-LCP                & 1.3514 $\pm$ 0.8619 & 0.0181 $\pm$ 0.0116 & \textbf{5.73 $\pm$ 4.31} & 45.06 $\pm$ 74.85 & 0.0373 $\pm$ 0.0355 \\
        SN $\lambda_{\text{SN}}=0.2$ (Ours) & \textbf{0.9083 $\pm$ 0.7559} & \textbf{0.0165 $\pm$ 0.0122} & 5.96 $\pm$ 9.40 & \textbf{29.33 $\pm$ 75.34} & 0.0550 $\pm$ 0.0870 \\
        \midrule
        \bottomrule
        \end{tabular}%
    }
    \label{tab:real_robot_results}
    \vspace{-10pt}
\end{table*}
 
\subsection{Effect of Spectral Normalization}
SN was applied to every layer of the actor network, and an additional scaling coefficient of $\lambda_{\text{SN}} = 0.2$ was multiplied on the final layer. The results, illustrated in \autoref{tab:ablation_studies}, indicate that SN leads to finite bandwidth action control, comparable to what is achieved by the GP-LCP approach.

\begin{figure}
    \centering
    \includegraphics[width=0.9\linewidth]{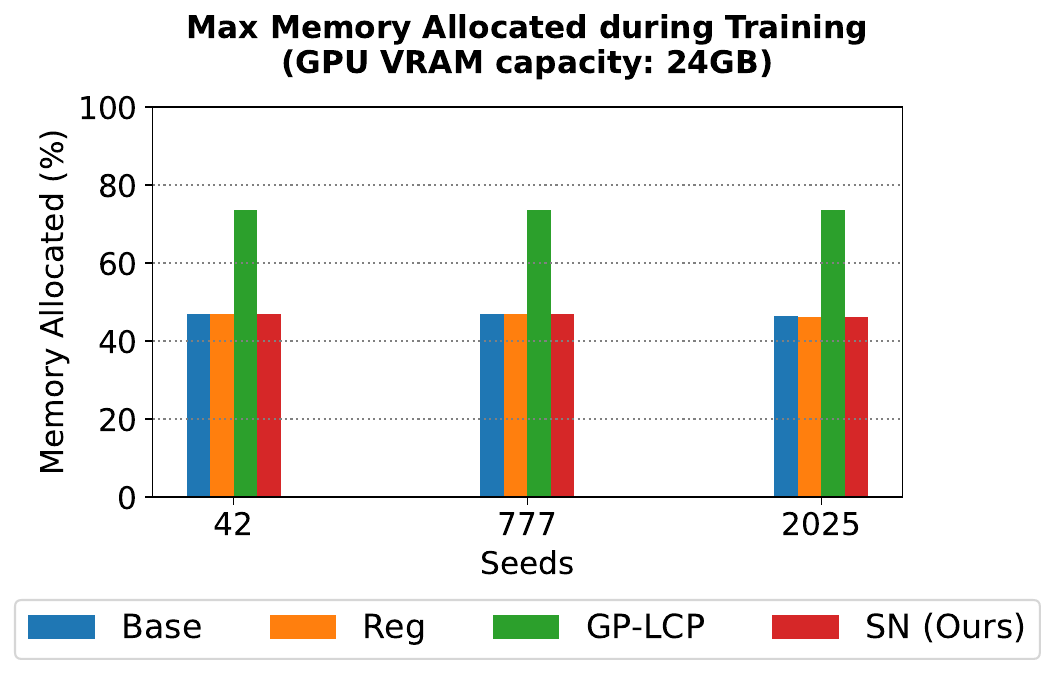}
    \caption{Maximum GPU memory allocation during training across different methods (Baseline, Reg (Regularization Reward), GP-LCP, and SN (Ours)) and multiple random seeds (42, 777, 2025). The results show that GP-LCP requires significantly more memory due to gradient penalty computations, while SN achieves comparable performance with lower memory overhead.}
    \label{fig:gpu}
    \vspace{-10pt}
\end{figure}

To highlight the memory efficiency of SN, experiments were performed on a NVIDIA GeForce RTX 4090 GPU, and the allocated memory usage was recorded. The GP-LCP method requires additional memory due to the overhead of its policy gradient calculations. \autoref{fig:gpu} illustrates that GP-LCP allocated approximately 74\% of the available GPU memory, which corresponds to around 18.98 GB, while SN only required 47\%, approximately 12.08 GB, a usage level similar to the baseline. This indicates that SN can match GP-LCP in terms of regularization and control quality with significantly lower memory usage.

\begin{figure}
    \centering
    \includegraphics[width=\linewidth]{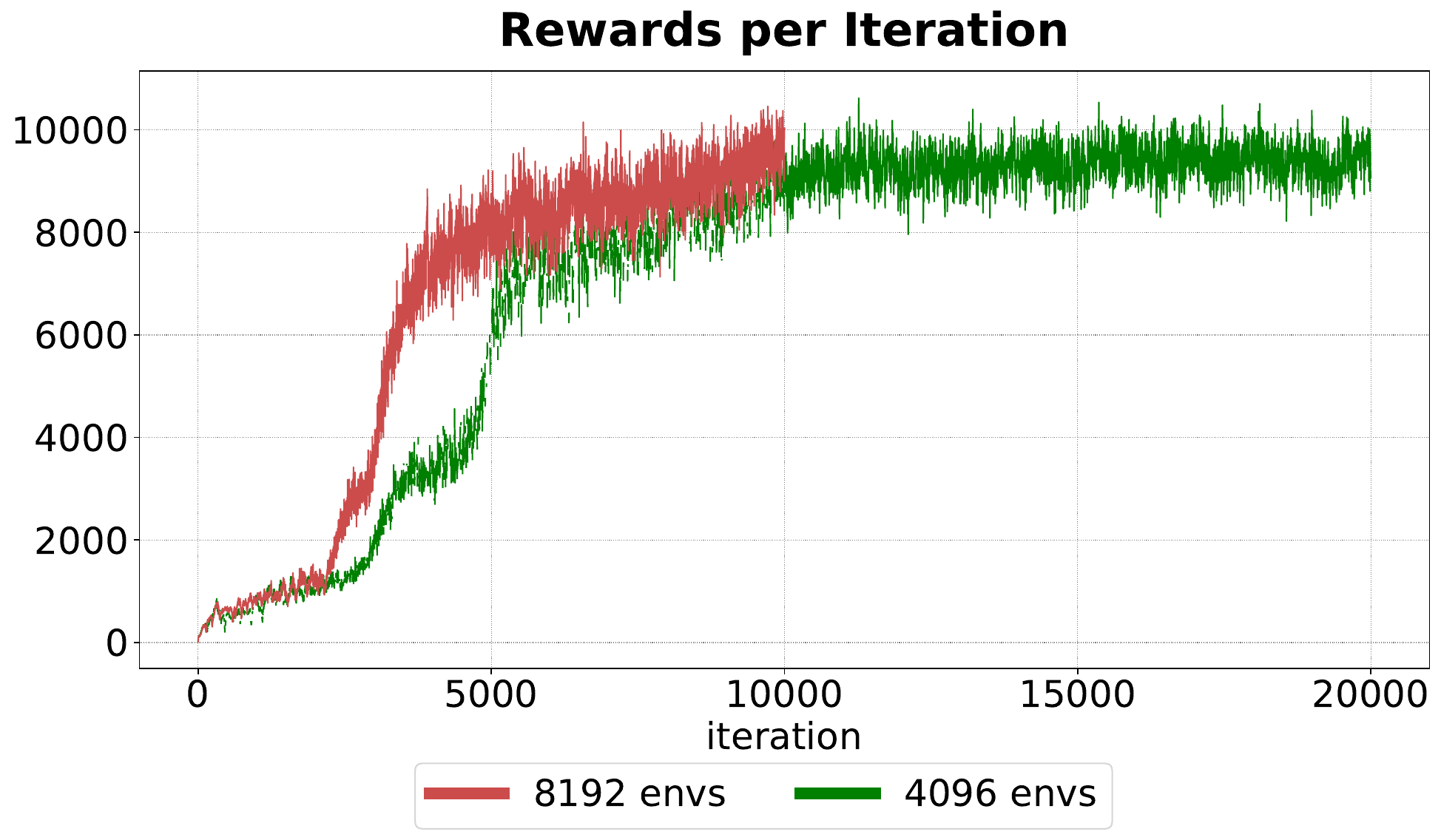}
    \caption{The plot compares training rewards per iteration for different numbers of environments (8192 vs. 4096) in simulation. The reduced GPU memory usage from SN, which eliminates the need for gradient penalty computations, allows for a higher number of parallel environments, leading to improved training efficiency.}
    \label{fig:num_envs}
    \vspace{-10pt}
\end{figure}

The memory efficiency of SN demonstrates that our approach can be trained on GPUs with relatively small memory capacities. Moreover, the memory savings can be leveraged to increase the batch size or the number of training environments, which in turn has the potential to accelerate convergence and enhance overall performance. \autoref{fig:num_envs} shows that the GPU memory savings from replacing GP-LCP with SN can be utilized to increase the number of parallel environments, leading to improved training efficiency. By utilizing the freed memory to scale up the number of environments, as demonstrated in \autoref{fig:num_envs}, SN enables faster data collection and more effective policy optimization.

\subsection{Effect of SN Coefficient \texorpdfstring{$\lambda_{\text{SN}}$}{lambda\_SN}}
\autoref{tab:ablation_studies}(b) presents a performance comparison for different values of the spectral normalization coefficient $\lambda_{\text{SN}}$. As indicated by \eqref{eq7}, the choice of $\lambda_{\text{SN}}$ affects the Lipschitz constant of the actor network. A higher $\lambda_{\text{SN}}=1.0$ leads to a looser upper bound on Lipschitz continuity, which can result in more jittery motions. Conversely, a very low $\lambda_{\text{SN}}$ can reduce exploration opportunities and slow down the learning process.

The best performance was observed at $\lambda_{\text{SN}} = 0.2$. When $\lambda_{\text{SN}}$ was increased to 1.0, motions became noticeably jittery, while a setting of 0.1 hindered learning to the point that the policy failed to converge. These results underscore the importance of carefully tuning $\lambda_{\text{SN}}$ to balance smoothness and exploration. 

\subsection{Real Robot Deployment}
To evaluate the effectiveness of each method on real hardware, the learned policies were implemented on our humanoid robot TOCABI and tested across three command inputs: \textit{forward walking}, \textit{rotational walking}, and \textit{walking forward while rotating}. Each test lasted 5 seconds, with a total of 1250 steps recorded for each scenario.

As shown in \autoref{tab:real_robot_results}, the baseline policy exhibited noticeable vibrations in all three walking scenarios, highlighting the necessity of action regularization for stable real-world deployment. The policy using the regularization reward also showed some minor oscillations during walking, indicating only partial mitigation of vibrations. This suggests that regularization rewards may not be highly effective in suppressing vibrations and can require extensive tuning to achieve stable behavior. Moreover, the persistent vibrations negatively impacted task return performance. In contrast, both the GP-LCP and SN policies demonstrated stable walking across all command inputs without any observable vibrations. Furthermore, while SN achieved the most stable regularization, its task return performance showed lower tracking accuracy. This confirms that SN can effectively enforce Lipschitz continuity with lower GPU memory usage, but also highlights the inherent trade-off between task performance and regularization. Therefore, careful hyperparameter tuning is required to achieve a well-balanced policy.

\section{CONCLUSIONS}
\label{sec:conclusions}
This work proposed the use of SN as an efficient approach for enforcing Lipschitz continuity in RL policies for humanoid locomotion. By constraining the spectral norm of network weights, SN effectively limits high-frequency action variations, achieving finite bandwidth control without the need for explicit gradient calculations. Experimental results demonstrated that SN achieves performance comparable to GP-LCP while significantly reducing GPU memory usage. The real-world deployment experiments further confirmed that SN-based policies result in stable walking behavior across various command inputs without inducing undesirable vibrations.

While SN provides a computationally efficient improvement over GP-LCP, this work has certain limitations. One limitation is that SN’s performance depends on the choice of the SN coefficient, requiring careful tuning to achieve optimal results across different tasks or environments. While hyperparameter tuning is a common requirement in regularization methods, selecting an appropriate SN coefficient remains an important consideration for balancing task performance and stability. Furthermore, although control metrics such as joint velocities, accelerations, and torque differences provided some insight into vibration levels, they were not always sufficient to precisely determine whether a policy would induce vibrations on real hardware. The final validation still required physical deployment, as certain hardware-specific factors affecting stability were difficult to capture purely through simulation.

Future research could explore adaptive SN coefficient strategies to automatically balance stability and performance, as well as investigate combining SN with other regularization methods to further enhance robustness in highly dynamic or unpredictable environments.

\addtolength{\textheight}{-12cm}   

\bibliographystyle{IEEEtran} 
\bibliography{IEEEabrv, IROS_ref}

\begin{thebibliography}{10}
\providecommand{\url}[1]{#1}
\csname url@samestyle\endcsname
\providecommand{\newblock}{\relax}
\providecommand{\bibinfo}[2]{#2}
\providecommand{\BIBentrySTDinterwordspacing}{\spaceskip=0pt\relax}
\providecommand{\BIBentryALTinterwordstretchfactor}{4}
\providecommand{\BIBentryALTinterwordspacing}{\spaceskip=\fontdimen2\font plus
\BIBentryALTinterwordstretchfactor\fontdimen3\font minus \fontdimen4\font\relax}
\providecommand{\BIBforeignlanguage}[2]{{%
\expandafter\ifx\csname l@#1\endcsname\relax
\typeout{** WARNING: IEEEtran.bst: No hyphenation pattern has been}%
\typeout{** loaded for the language `#1'. Using the pattern for}%
\typeout{** the default language instead.}%
\else
\language=\csname l@#1\endcsname
\fi
#2}}
\providecommand{\BIBdecl}{\relax}
\BIBdecl

\bibitem{siekmann2021sim}
J.~Siekmann, Y.~Godse, A.~Fern, and J.~Hurst, ``Sim-to-real learning of all common bipedal gaits via periodic reward composition,'' in \emph{2021 IEEE International Conference on Robotics and Automation (ICRA)}.\hskip 1em plus 0.5em minus 0.4em\relax IEEE, 2021, pp. 7309--7315.

\bibitem{duan2022sim}
H.~Duan, A.~Malik, J.~Dao, A.~Saxena, K.~Green, J.~Siekmann, A.~Fern, and J.~Hurst, ``Sim-to-real learning of footstep-constrained bipedal dynamic walking,'' in \emph{2022 International Conference on Robotics and Automation (ICRA)}.\hskip 1em plus 0.5em minus 0.4em\relax IEEE, 2022, pp. 10\,428--10\,434.

\bibitem{kim_tbasedrl}
D.~Kim, G.~Berseth, M.~Schwartz, and J.~Park, ``Torque-based deep reinforcement learning for task-and-robot agnostic learning on bipedal robots using sim-to-real transfer,'' \emph{IEEE Robotics and Automation Letters}, vol.~8, no.~10, pp. 6251--6258, 2023.

\bibitem{tang2024humanmimic}
A.~Tang, T.~Hiraoka, N.~Hiraoka, F.~Shi, K.~Kawaharazuka, K.~Kojima, K.~Okada, and M.~Inaba, ``Humanmimic: Learning natural locomotion and transitions for humanoid robot via wasserstein adversarial imitation,'' in \emph{2024 IEEE International Conference on Robotics and Automation (ICRA)}.\hskip 1em plus 0.5em minus 0.4em\relax IEEE, 2024, pp. 13\,107--13\,114.

\bibitem{kim2024not}
Y.~Kim, H.~Oh, J.~Lee, J.~Choi, G.~Ji, M.~Jung, D.~Youm, and J.~Hwangbo, ``Not only rewards but also constraints: Applications on legged robot locomotion,'' \emph{IEEE Transactions on Robotics}, 2024.

\bibitem{kim2024learning}
G.~Kim, Y.-H. Lee, and H.-W. Park, ``A learning framework for diverse legged robot locomotion using barrier-based style rewards,'' \emph{arXiv preprint arXiv:2409.15780}, 2024.

\bibitem{fu2021minimizing}
Z.~Fu, A.~Kumar, J.~Malik, and D.~Pathak, ``Minimizing energy consumption leads to the emergence of gaits in legged robots,'' \emph{arXiv preprint arXiv:2111.01674}, 2021.

\bibitem{chen2024lcp}
Z.~Chen, X.~He, Y.-J. Wang, Q.~Liao, Y.~Ze, Z.~Li, S.~S. Sastry, J.~Wu, K.~Sreenath, S.~Gupta, and X.~B. Peng, ``Learning smooth humanoid locomotion through lipschitz-constrained policies,'' \emph{arxiv preprint arXiv:2410.11825}, 2024.

\bibitem{makoviychuk2021isaac}
V.~Makoviychuk, L.~Wawrzyniak, Y.~Guo, M.~Lu, K.~Storey, M.~Macklin, D.~Hoeller, N.~Rudin, A.~Allshire, A.~Handa, and G.~State, ``Isaac gym: High performance gpu-based physics simulation for robot learning,'' 2021.

\bibitem{cheng2024extreme}
X.~Cheng, K.~Shi, A.~Agarwal, and D.~Pathak, ``Extreme parkour with legged robots,'' in \emph{2024 IEEE International Conference on Robotics and Automation (ICRA)}.\hskip 1em plus 0.5em minus 0.4em\relax IEEE, 2024, pp. 11\,443--11\,450.

\bibitem{kumar2021rma}
A.~Kumar, Z.~Fu, D.~Pathak, and J.~Malik, ``Rma: Rapid motor adaptation for legged robots,'' \emph{arXiv preprint arXiv:2107.04034}, 2021.

\bibitem{lai2023sim}
H.~Lai, W.~Zhang, X.~He, C.~Yu, Z.~Tian, Y.~Yu, and J.~Wang, ``Sim-to-real transfer for quadrupedal locomotion via terrain transformer,'' in \emph{2023 IEEE International Conference on Robotics and Automation (ICRA)}.\hskip 1em plus 0.5em minus 0.4em\relax IEEE, 2023, pp. 5141--5147.

\bibitem{kumar2022adapting}
A.~Kumar, Z.~Li, J.~Zeng, D.~Pathak, K.~Sreenath, and J.~Malik, ``Adapting rapid motor adaptation for bipedal robots,'' in \emph{2022 IEEE/RSJ International Conference on Intelligent Robots and Systems (IROS)}.\hskip 1em plus 0.5em minus 0.4em\relax IEEE, 2022, pp. 1161--1168.

\bibitem{radosavovic2024real}
I.~Radosavovic, T.~Xiao, B.~Zhang, T.~Darrell, J.~Malik, and K.~Sreenath, ``Real-world humanoid locomotion with reinforcement learning,'' \emph{Science Robotics}, vol.~9, no.~89, p. eadi9579, 2024.

\bibitem{he2024learning}
T.~He, Z.~Luo, W.~Xiao, C.~Zhang, K.~Kitani, C.~Liu, and G.~Shi, ``Learning human-to-humanoid real-time whole-body teleoperation,'' in \emph{2024 IEEE/RSJ International Conference on Intelligent Robots and Systems (IROS)}.\hskip 1em plus 0.5em minus 0.4em\relax IEEE, 2024, pp. 8944--8951.

\bibitem{he2024omnih2o}
T.~He, Z.~Luo, X.~He, W.~Xiao, C.~Zhang, W.~Zhang, K.~Kitani, C.~Liu, and G.~Shi, ``Omnih2o: Universal and dexterous human-to-humanoid whole-body teleoperation and learning,'' \emph{arXiv preprint arXiv:2406.08858}, 2024.

\bibitem{fu2024humanplus}
Z.~Fu, Q.~Zhao, Q.~Wu, G.~Wetzstein, and C.~Finn, ``Humanplus: Humanoid shadowing and imitation from humans,'' \emph{arXiv preprint arXiv:2406.10454}, 2024.

\bibitem{liu2024visual}
M.~Liu, Z.~Chen, X.~Cheng, Y.~Ji, R.-Z. Qiu, R.~Yang, and X.~Wang, ``Visual whole-body control for legged loco-manipulation,'' \emph{arXiv preprint arXiv:2403.16967}, 2024.

\bibitem{gu2024humanoid}
X.~Gu, Y.-J. Wang, and J.~Chen, ``Humanoid-gym: Reinforcement learning for humanoid robot with zero-shot sim2real transfer,'' \emph{arXiv preprint arXiv:2404.05695}, 2024.

\bibitem{snforgan}
\BIBentryALTinterwordspacing
T.~Miyato, T.~Kataoka, M.~Koyama, and Y.~Yoshida, ``Spectral normalization for generative adversarial networks,'' \emph{CoRR}, vol. abs/1802.05957, 2018. [Online]. Available: \url{http://arxiv.org/abs/1802.05957}
\BIBentrySTDinterwordspacing

\bibitem{o2006metric}
M.~O'Searcoid, \emph{Metric spaces}.\hskip 1em plus 0.5em minus 0.4em\relax Springer Science \& Business Media, 2006.

\bibitem{schwartz2022design}
M.~Schwartz, J.~Sim, J.~Ahn, S.~Hwang, Y.~Lee, and J.~Park, ``Design of the humanoid robot tocabi,'' in \emph{2022 IEEE-RAS 21st International Conference on Humanoid Robots (Humanoids)}.\hskip 1em plus 0.5em minus 0.4em\relax IEEE, 2022, pp. 322--329.

\bibitem{yoshida2017spectral}
Y.~Yoshida and T.~Miyato, ``Spectral norm regularization for improving the generalizability of deep learning,'' \emph{arXiv preprint arXiv:1705.10941}, 2017.

\bibitem{peng2021amp}
X.~B. Peng, Z.~Ma, P.~Abbeel, S.~Levine, and A.~Kanazawa, ``Amp: Adversarial motion priors for stylized physics-based character control,'' \emph{ACM Transactions on Graphics (ToG)}, vol.~40, no.~4, pp. 1--20, 2021.

\bibitem{schulman2017proximal}
J.~Schulman, F.~Wolski, P.~Dhariwal, A.~Radford, and O.~Klimov, ``Proximal policy optimization algorithms,'' \emph{arXiv preprint arXiv:1707.06347}, 2017.

\end{thebibliography}

\end{document}